\documentclass[letterpaper, 10 pt, conference]{ieeeconf}  

\IEEEoverridecommandlockouts                              

\usepackage{graphics} 
\usepackage{epsfig} 
\usepackage{mathptmx} 
\usepackage{times} 
\usepackage{amsmath} 
\usepackage{amssymb}  
\usepackage[dvipsnames]{xcolor}
\usepackage[hidelinks]{hyperref} 

\definecolor{myYellow}{RGB}{230, 180, 0}

\title{\LARGE \bf
Self-Supervised Enhancement for Depth from a Lightweight ToF Sensor with Monocular Images 
}

\author{Laiyan Ding$^{1}$, Hualie Jiang$^{2}$, Jiwei Chen$^{1}$, and Rui Huang$^{1*}$
\thanks{$^{*}$Corresponding author}
\thanks{$^{1}$Laiyan Ding, Jiwei Chen and Rui Huang are with School of Science and Engineering, The Chinese University of Hong Kong, Shenzhen 518172, China (e-mail: laiyanding@link.cuhk.edu.cn; jiweichen2@link.cuhk.edu.cn; ruihuang@cuhk.edu.cn).}
\thanks{$^{2}$Hualie Jiang is with Insta360 Research, Shenzhen 518000, China (e-mail: jianghualie@insta360.com).}}

\begin{document}

\maketitle
\thispagestyle{empty}
\pagestyle{empty}

\begin{abstract}


Depth map enhancement using paired high-resolution RGB images offers a cost-effective solution for improving low-resolution depth data from lightweight ToF sensors. Nevertheless, naively adopting a depth estimation pipeline to fuse the two modalities requires groundtruth depth maps for supervision. To address this, we propose a self-supervised learning framework, SelfToF, which generates detailed and scale-aware depth maps. Starting from an image-based self-supervised depth estimation pipeline, we add low-resolution depth as inputs, design a new depth consistency loss, propose a scale-recovery module, and finally obtain a large performance boost. Furthermore, since the ToF signal sparsity varies in real-world applications, we upgrade SelfToF to SelfToF* with submanifold convolution and guided feature fusion. Consequently, SelfToF* maintain robust performance across varying sparsity levels in ToF data. Overall, our proposed method is both efficient and effective, as verified by extensive experiments on the NYU and ScanNet datasets. The code is available at \href{https://github.com/denyingmxd/selftof}{https://github.com/denyingmxd/selftof}.



\end{abstract}

\section{INTRODUCTION}

Accurate depth perception is an essential component to a wide range of applications, including visual semantic navigation \cite{liang2021sscnav}, holistic scene understanding \cite{wang2024embodiedscan}, \textit{etc.}. As a result, depth sensors, e.g., Lidar and Time-of-Flight (ToF) cameras, have gained much attention due to their precision and robustness. Nevertheless, they have high prices and energy consumption (e.g., Intel RealSense D435i costs around 300\$ and runs at 7W). As an alternative, lightweight ToF sensors (e.g., ST VL53L5CX \cite{VL53L5CX}, dubbed L5) are low-cost and have been used widely in factories for IoT devices. Specifically, L5 only costs less than 5\$ and operates at 0.3w. As a pioneer work, DELTAR \cite{li2022deltar} have made enhancing the depth map from a L5 with RGB images possible. Depth maps from DELTAR can be further used for other tasks, including Nerf-based SLAM \cite{liu2023multi}. Note that we regard this task as depth enhancement as it differs from depth superresolution \cite{wang2024sgnet,he2021towards} and depth completion \cite{park2020non,zhang2023completionformer}, in terms of inputs which have extremely low resolution and potential missing zones. Unfortunately, DELTAR still need dense depth groundtruth for supervision, which limits its applications. 


\begin{figure}[t]
    \centering
    \includegraphics[width=0.5\textwidth]{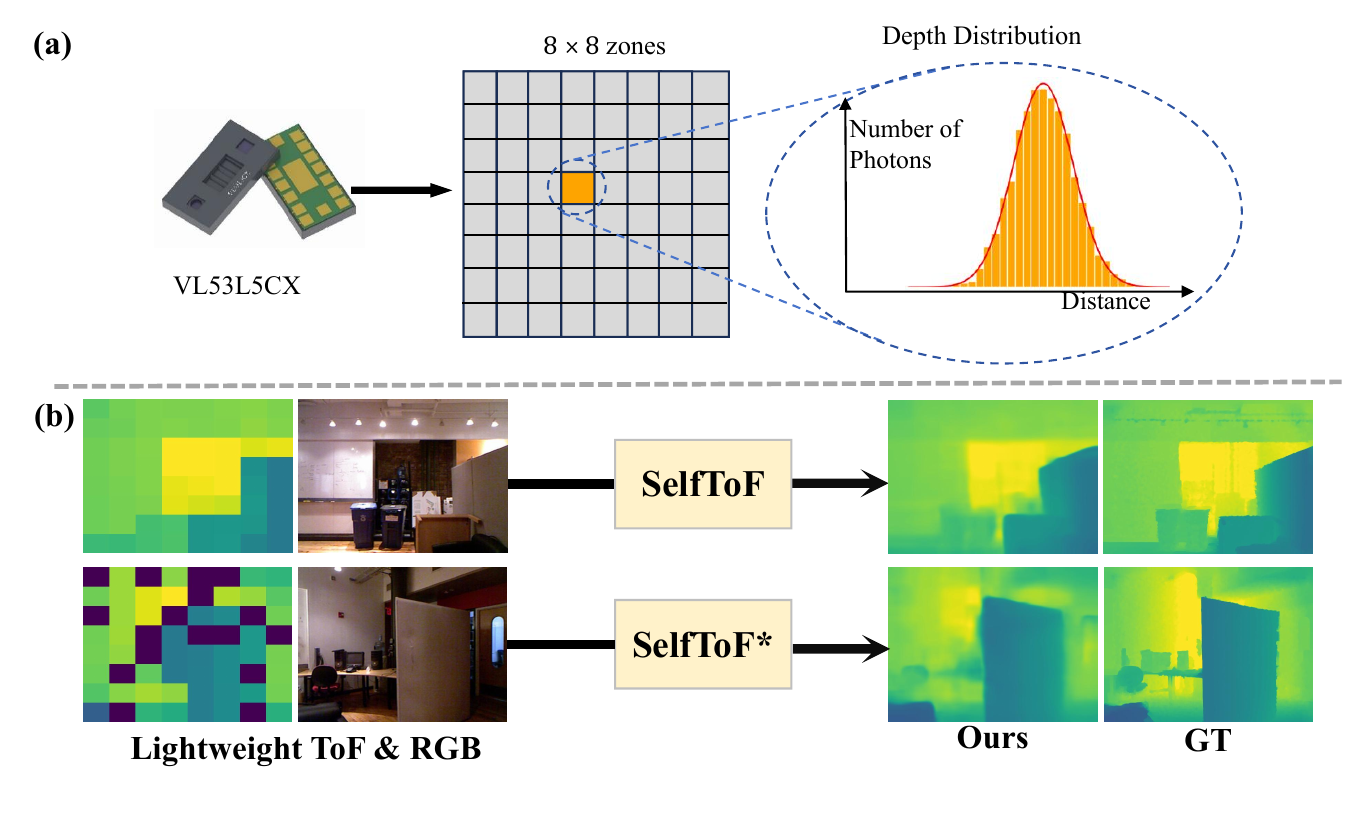}
    \caption{L5 sensing principle and example inputs and outputs with our SelfToF and SelfToF*. (a) L5 would return zones of resolution $8 \times 8$, and each zone provides depth distribution information. (b) SelfToF and SelfToF* can generate reasonable depth maps in different sparsity setups.}
    \label{fig:intro}
\end{figure}


To eliminate the burden of collecting groundtruth depth maps for training, we take inspiration from the self-supervised depth estimation methods, e.g., Monodepth2 \cite{godard2019digging}. These methods can predict depth maps and poses (with the inherent DepthNet and PoseNet) from monocular images without groundtruth supervision. Consequently, we build our SelfToF starting from Monodepth2. We first feed input depth to the DepthNet in addition to the RGB image and design a zone-wise depth consistency loss, as in self-supervised depth completion \cite{ma2019self}. Nevertheless, the weight of this depth consistency loss cannot be set high as in previous works due to its imprecise constraint, and the output depth map is still scale-ambiguous. Thus, we introduce a median of median scaling method (MMS), which exploits the zone-wise nature of input depth to recover the metric scale. Consequently, our model is metric-aware and do not need further scaling during testing. Additionally, we propose to feed the input depth to the PoseNet for better pose and depth estimation. Overall, these ingredients allow a large performance boost.  

To cope with invalid or missing zones (similar to different sparsity levels in depth completion \cite{long2021depth}), we introduce SelfToF*, which replaces the depth encoder in SelfToF with a submanifold one and adds a guided feature fusion module to mitigate performance drop due to the sparsity. This way, the propagation of the input depth feature is carefully controlled and leads to the robustness of SelfToF*. Example inputs and outputs of SelfToF and SelfToF* are in Fig. \ref{fig:intro} (b). Compared with the baseline Monodepth2, SelfToF reduce abs\_rel from 0.154 to 0.049. Furthermore, SelfToF* retain the abs\_rel of 0.053 while SelfToF rise to 0.060.

Our contributions can be summarized as follows. Firstly, we propose SelfToF, a self-supervised framework that can enhance the depth map from a lightweight ToF sensor with paired images. Secondly, quantitative and qualitative results verify that SelfToF performs better than previous architectures designed for self-supervised estimation \cite{godard2019digging,jiang2021plnet}, or guided depth upsampling \cite{he2012guided}. Lastly, we introduce a submanifold depth encoder and guided feature fusion to upgrade from SelfToF to SelfToF*, which is more robust to sparsity variation from the lightweight ToF sensor.


\section{Related Works}

\subsection{Self-Supervised Monocular Depth Estimation} 

These approaches do not require depth ground truth \cite{fu2018deep} or rectified stereo pairs \cite{garg2016unsupervised} for end-to-end depth learning. Instead, they learn the depth and inter-frame poses simultaneously \cite{zhou2018unsupervised}. The supervisory signal originates from the assumed photometric consistency between the given current frame and the reconstructed current frame from temporally adjacent frames. Thus, the depth and pose networks can learn end-to-end with only monocular images. Later on, great efforts including devising heuristic loss functions \cite{bian2019unsupervised,godard2019digging,zhou2018unsupervised}, improving robustness to moving objects \cite{jiang2021unsupervised,casser2019depth}, extending to modern surround applications \cite{kim2022self,ding2024towards,wei2023surrounddepth} and designing efficient networks \cite{karpov2022exploring,ramamonjisoa2021single}. These works mostly focus on problems in outdoor and autonomous driving scenarios, while another line of research pays attention to challenges rooted in indoor scenes. The progress that tackles textureless regions and large rotations has been made \cite{jiang2021plnet,zhou2019moving,yu2020p,bian2021auto,zhao2023gasmono}. More importantly, these methods based on monocular images can predict depth maps with high resolution and rich details. Thus, we aim to enhance the low-resolution depth map from a lightweight ToF sensor with this self-supervised learning paradigm.

\subsection{Self-Supervised Monocular Depth Completion} 

Ma \textit{et al.} propose a self-supervised depth completion framework \cite{ma2019self} that requires only sparse depth and paired RGB images as inputs and supervision during training or testing. Later, VOICED \cite{wong2020unsupervised} explores how to use SLAM systems with self-supervised depth completion. KBNet \cite{wong2021unsupervised} design a sparse-to-dense module that densifies sparse depth input and a Calibrated Backprojection Layer which can inject spatial positional encoding into the network. SelfDeco \cite{choi2021selfdeco} employs a deep stack of sparse convolution layers \cite{uhrig2017sparsity} and pixel adaptive convolution \cite{su2019pixel} to fuse features from different modalities. Mondi \cite{liu2022monitored} investigates how to effectively learn from a blind ensemble of teacher models by using photometric reconstruction errors as the criterion. Differently, our model takes a low-resolution depth map as input rather than a sparse depth map. Moreover, we introduce a submanifold depth encoder and a guided feature fusion to mitigate performance drop in different ToF sparsity levels.

\subsection{Lightweight-ToF-based Depth Estimation}

DELTAR \cite{li2022deltar} predicts a high-resolution depth map given paired RGB image and low-resolution depth information from a lightweight ToF sensor. The produced depth map can help develop the SLAM system \cite{liu2023multi} which is based on NeRF \cite{mildenhall2021nerf}. Ours is a self-supervised framework that can be trained end-to-end without groundtruth depth as supervision. This advantage of our system allows more potential applications, similar to self-supervised depth estimation \cite{godard2019digging,zhou2018unsupervised}. 

\section{Method}



Similar to self-supervised depth estimation \cite{godard2019digging}, our task of self-supervised enhancement for a depth map from a lightweight ToF sensor with images requires a DepthNet and a PoseNet. Both methods are trained on multiple temporally adjacent frames without depth groundtruth supervision. Differently, our model takes additional paired depth maps from lightweight ToF sensors as inputs and produces scale-aware depth output rather than scale-ambiguous ones \cite{zhou2018unsupervised}.  

In the following, we first briefly introduce the sensing mechanism of the lightweight ToF sensor, e.g., L5 \cite{VL53L5CX}. Then, we illustrate the overall framework of our SelfToF, which takes inspiration from self-supervised depth estimation and completion \cite{ma2019self}. Lastly, we describe our additional designs containing submanifold depth encoder and guided feature fusion, which can mitigate the performance drop of our model given different sparsities of ToF signals, i.e., zeros in the input low-resolution depth map.

\subsection{Sensing Mechanism of L5}
As mentioned above, L5 is low-cost and low-energy compared with traditional depth sensors, e.g., RealSense and Kinect V2. Nevertheless, it has a very low resolution of only $8 \times 8$ (64 zones in total), and each zone measures the depth distribution of its corresponding scene in the 3D scene. Specifically, as depicted in Fig. \ref{fig:intro} (a), the L5 sensor evaluates depth by counting photons received within specific time frames. This data is then approximated using a Gaussian distribution, allowing the sensor to transmit only the mean and variance \cite{li2022deltar}. In this work, we would take these Gaussians' mean as inputs, forming a depth map of resolution $8 \times 8$.

\begin{figure}[t]
    \centering
    \includegraphics[width=0.5\textwidth]{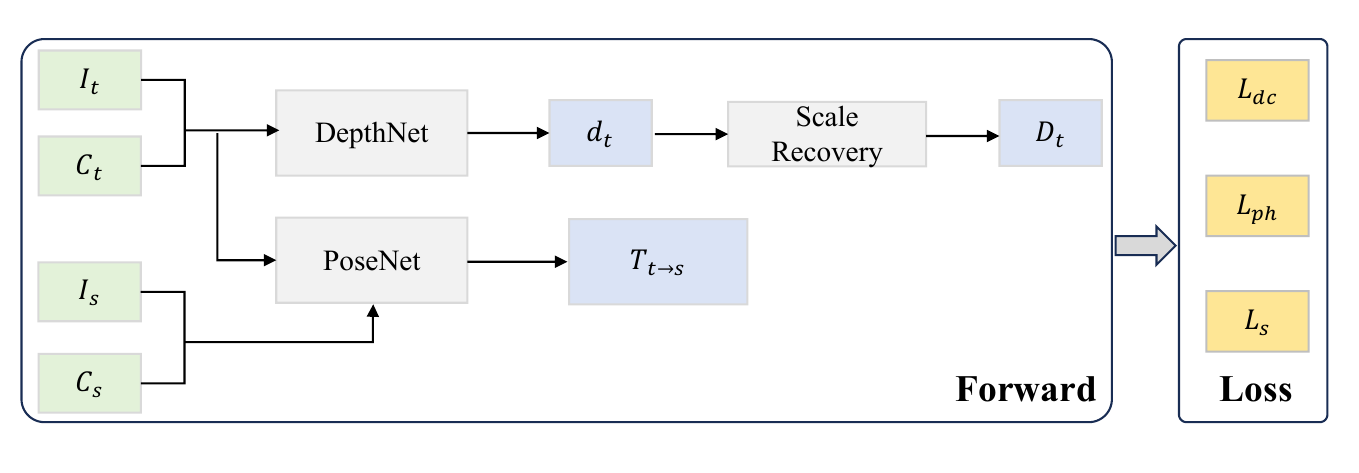}
    \caption{Training pipeline of our model. During forward, the model takes paired RGB images and low-resolution depth maps from L5 as inputs and outputs depth maps for the current frame and the relative pose between the current frame and source frame. Then, the loss functions, including depth consistency loss ($L_{dc}$), photometric loss ($L_{ph}$), and smoothness loss ($L_s$), can be computed to supervise the DepthNet and PoseNet jointly.}
    \label{fig:network}
\end{figure}

\subsection{Overall Framework}


Assuming that L5 zones are aligned with RGB images and no maximum distance constraint exists, then the training pipeline of SelfToF is illustrated in Fig. \ref{fig:network}. SelfToF takes low-resolution depth maps ($C_t, C_s$) from a lightweight ToF sensor and paired RGB images ($I_t, I_s$) as inputs, where $t$ and $s$ are the indexes of the target frame and the source frame. 
The DepthNet can generate a scale-ambiguous depth output $d_t$, 
which can be further transformed to metric-aware $D_t$ through our scale-recovery module. 
The PoseNet takes paired images and depth maps as inputs and outputs the relative pose $T_{t \to s}$ between the target frame and the source frame. 
Lastly, we calculate the overall loss function as the weighted summation among the depth consistency loss ($L_{dc}$), photometric loss ($L_{ph}$),  and smoothness loss ($L_s$). Note that, during testing, only DepthNet is required to perform depth enhancement.


\subsubsection{DepthNet}
The DepthNet consists of an RGB encoder, a depth encoder, a fusion module, and a decoder, as in Fig. \ref{fig:sub_networks} (a). The RGB encoder is a ResNet18 \cite{he2016deep} initialized from ImageNet \cite{deng2009imagenet} pretrained weights. The depth encoder is a simple network that contains multiple $1 \times1$ and $3 \times 3$ convolution layers and operates on the $8 \times 8$ resolution without downsampling. The RGB encoder outputs feature maps of different resolutions, while ToF encoders produce the same number of feature maps of the same spatial size. Then paired feature maps are fused (addition or guided feature fusion in Sec. \ref{sec:guide}) and passed to the lightweight decoder, which has consecutive convolution and upsampling operations, as in Monodepth2. Note that if we use addition as the fusion between depth and RGB features, compared with Monodepth2, the Multiply-Accumulate Operations (MACs) only increase from 2.38G to 2.59 G. Consequently, SelfToF is efficient as it can run at more than 100 FPS with a single RTX2080Ti GPU with a RGB resolution of $256\times256$.

\subsubsection{PoseNet}
\begin{figure}[t]
    \centering
    \includegraphics[width=0.5\textwidth]{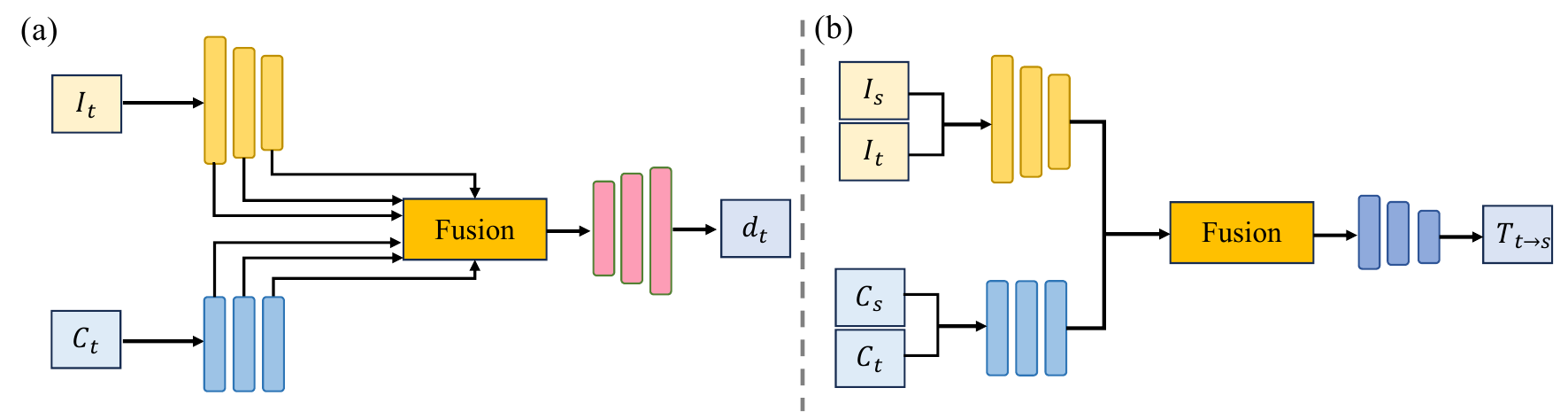}
    \caption{Overview of multi-modality DepthNet and PoseNet. The fusion is a simple addition fusion and our guided feature fusion in SelfToF and SelfToF*, respectively. }
    \label{fig:sub_networks}
\end{figure}

The PoseNet also includes an RGB encoder, a depth encoder, a fusion module, and a decoder, as in Fig. \ref{fig:sub_networks} (b). The process is similar to DepthNet, except only the feature maps at the lowest level are fused and decoded for rotation and translation. Though previous works \cite{ma2019self,wong2021unsupervised,choi2021selfdeco} have tried different ways to estimate the relative temporal pose, they only use the RGB information. Differently, we utilize the input depth information for better pose estimation. This small modification allows a large performance boost as validated by Table \ref{table:ideal_ablate}.

\subsubsection{Scale Recovery}
\label{sec: scale}
Following prior self-supervised depth completion works, we design a depth consistency loss $L_{dc}$ (Eq. \ref{eq:dc}) to maintain the consistency of output and input depth. Differently, $L_{dc}$ is a zone-wise loss function instead of a point-wise loss. Consequently, we notice that assigning a large weight to $L_{dc}$ (e.g., 1.0) as previous works \cite{ma2019self} will force the output to be too similar to the input depth. Furthermore, the other two loss functions, $L_{ph}$ and $L_s$, cannot help reveal the metric scale. Thus, simply lowering the weight of $L_{dc}$ cannot guarantee that the output depth is scale-aware. To this end, we propose a scale-recovery module to recover the actual scale without increasing the weight of $L_{dc}$. 

A naive way is to conduct median scaling similar to the median scaling (MS) in the self-supervised depth estimation evaluation protocol \cite{zhou2018unsupervised}. That is, 
\begin{equation}
D_t = \frac{median(C_t)}{median(d_t)} * {d_t}
\end{equation}

This is indeed useful but does not consider the zone-wise nature of ToF signals. Thus, we can first obtain the zone-wise scales and then take the median of these scales as the final overall scale. That is,

\begin{equation}
D_t = median(
\left\{\frac{C_t^i}{median(d_t^i)}
\right\}
)
* {d_t}
\end{equation}
where $C_t^i$ and $d_t^i$ are the ith zone in $C_t$ and $d_t$. Moreover, we name this as the median of median scaling (MMS).

\subsection{Loss Functions}

\subsubsection{Photometric Loss}

With the camera intrinsic matrix $K$, the relative pose $T_{t \rightarrow s}$, and the depth of the current frame $D_t$, we can synthesize the current frame $I_{s \rightarrow t}$ from the source frame with backward warping. Same as previous works, this synthesis follows:
\begin{equation}
I_{s \rightarrow t}(x) = I_s \langle proj(D_t, K,  T_{t \to s}) \rangle.
\end{equation}
where $proj()$ is calculates the resulting 2D coordinates in the source image, $\langle \rangle$ is the bilinear sampling function \cite{jaderberg2015spatial} for non-integer pixel coordinates.

Then the photometric loss $L_{ph}$ is the weighted sum of the L1 loss and the SSIM loss \cite{wang2004image}:
\begin{equation}
L_{ph}(I_t, I_{s \rightarrow t}) = \alpha \frac{1 - \text{SSIM}(I_t, I_{s \rightarrow t})}{2} + (1 - \alpha) \left\| I_t - I_{s \rightarrow t} \right\|,
\end{equation}
where $\alpha$ is conventionally set to be 0.85.

\subsubsection{Smoothness Loss}
Following previous works, we also include a smoothness loss $L_{s}$ to exploit the textures from the RGB image for depth estimation:
\begin{equation}
L_{s} = \left| \partial_x d_t^* \right| e^{-|\partial_x I_t|} + \left| \partial_y d_t^* \right| e^{-|\partial_y I_t|},
\end{equation}
where $\partial_x$ and $\partial_y$ represent the gradients along the x and y direction. $d_t^*$ is the mean normalized disparity \cite{wang2018learning} to prevent the shrinking of predicted disparity.

\subsubsection{Depth Consistency Loss}
Since a lightweight ToF sensor can provide the mean $\mu_i$ and standard deviation $\sigma_i$ of fitted Gaussian distribution of the histogram for the $ith$ zone \cite{li2022deltar}, we conduct this procedure for the predicted depth map to get $\Tilde{\mu_i}$ and $\Tilde{\sigma_i}$. Then our depth consistency loss $L_{dc}$ is defined as the distance between two 1D Gaussian distributions \cite{givens1984class}:
\begin{equation}
L_{dc} = \sum_i{(\mu_i-\Tilde{\mu_i})^2 + (\sigma_i-\Tilde{\sigma_i})^2},
\label{eq:dc}
\end{equation}
Notice that this equation can still be applied without the Gaussian assumption since it penalizes the differences between the two sets of mean and standard deviation.

\subsubsection{Overall Loss Function}

Our final loss for training is:

\begin{equation}
L = w_{ph}L_{ph} + w_{s}L_s + w_{dc}L_{dc},
\label{eq:final}
\end{equation}
where $w_{ph}$, $w_s$, and $w_{dc}$ are empirically set as 1.0, 0.1, and 0.01.

\subsection{Techniques to tackle with Sparsity in ToF}
Similar to prior studies on depth completion \cite{zhang2023completionformer,wong2021unsupervised}, we conduct experiments where the sparsity of ToF varies by setting some of the zones as invalid or missing and filling them with zeros. As expected, the depth prediction performance in both valid and invalid zones suffers from drops. We argue that during depth feature encoding, valid zones may gather useless information from invalid zones and invalid zones cannot gather most related features from valid zones correctly. Thus, proper control over the depth feature propagation is essential to mitigate the performance drop due to invalid zones. To this end, we introduce a submanifold depth encoder and a guided feature fusion module to upgrade SelfToF to SelfToF*, which suffers from less performance drop when some zones are invalid in the input depth map. 

\subsubsection{Submanifold Depth Encoder}

We turn the conventional depth feature encoder into a submanifold depth encoder by replacing all layers with their submanifold versions \cite{SubmanifoldSparseConvNet} implemented by Spconv \cite{spconv2022}. This way, the valid zones will gather information from only valid zones, and invalid zones will remain zeros. Different from previous works that use sparsity invariant convolution layers \cite{choi2021selfdeco, uhrig2017sparsity} for better performance, we use submanifold layers \cite{SubmanifoldSparseConvNet} for preventing 
uncontrolled depth feature propagation. Also, we validate that this submanifold encoder is necessary for guided feature fusion to be effective in Table \ref{table:ablate_missing}.

\subsubsection{Guided Feature Fusion}
\label{sec:guide}
When there is no sparsity in ToF, we use addition to fuse RGB and depth features. Nevertheless, when the sparsity exists, we aim to first propagate depth features under the guidance of RGB features and pixel positions. This allows the invalid zones to collect most related features from other zones and better fusion. To realize this intuition, we take inspiration from affinity-based methods \cite{frey2007clustering,ke2018adaptive} and self-attention \cite{vaswani2017attention}.

\begin{figure}[t]
    \centering
    \includegraphics[width=0.5\textwidth]{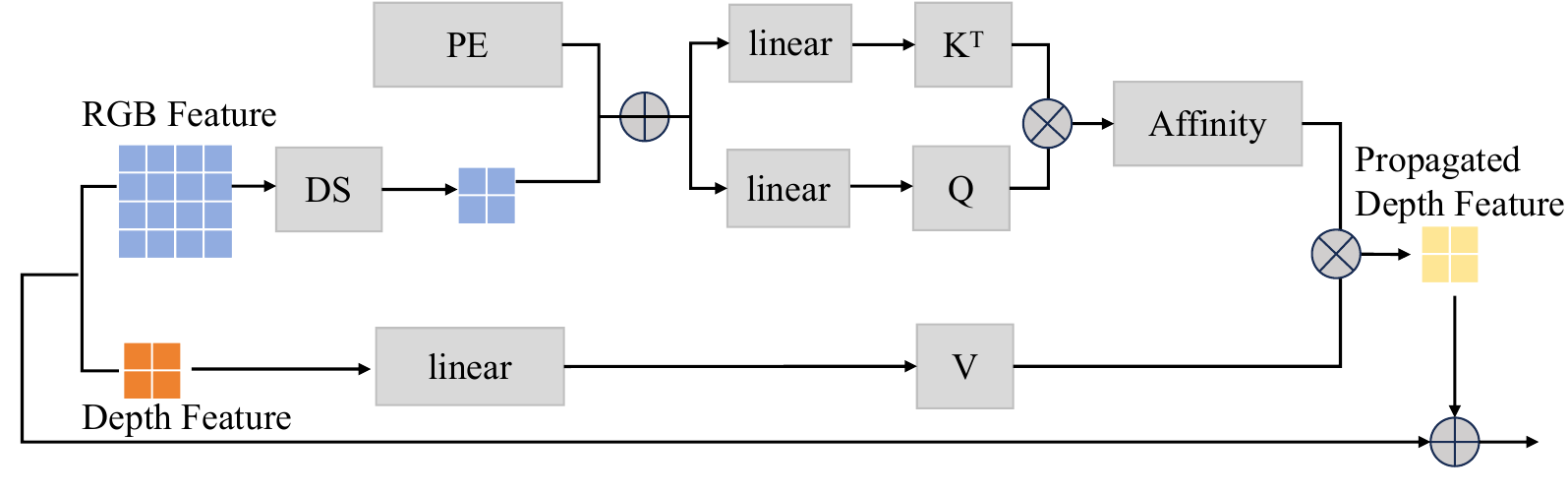}
    \caption{Overview of guided feature fusion. We first downsample (DS) RGB features and add position encoding \cite{vaswani2017attention} to it. Then, an affinity map is calculated to help propagate ToF features. Finally, the output of this module is the summation of the original RGB features, depth features, and propagated depth features. Resizing and flattening are omitted for simplicity.}
    \label{fig:propagation}
\end{figure}

Specifically speaking, as illustrated in Fig. \ref{fig:propagation}, given RGB and depth features, we first downsample the RGB feature to the resolution of the depth feature and flatten both the low-resolution RGB and depth features. Then, after the position embedding \cite{vaswani2017attention} is added to the low-resolution RGB features, two linear layers are responsible for generating the key ($K^T$), which is transposed, and value ($Q$) tensors. With a multiplication between the transposed key and value, we obtain the affinity matrix that is both semantic-aware and position-aware. Next, this affinity matrix propagates the generated value tensor from depth features through another matrix multiplication. Lastly, this module outputs the summation of the RGB feature, the propagated depth feature, and the original depth feature. Since this module operates on the $8 \times 8$ resolution, it brings little computation burden to the whole network.

\section{Experiments}
\begin{figure}[t]
    \centering
    \includegraphics[width=0.5\textwidth]{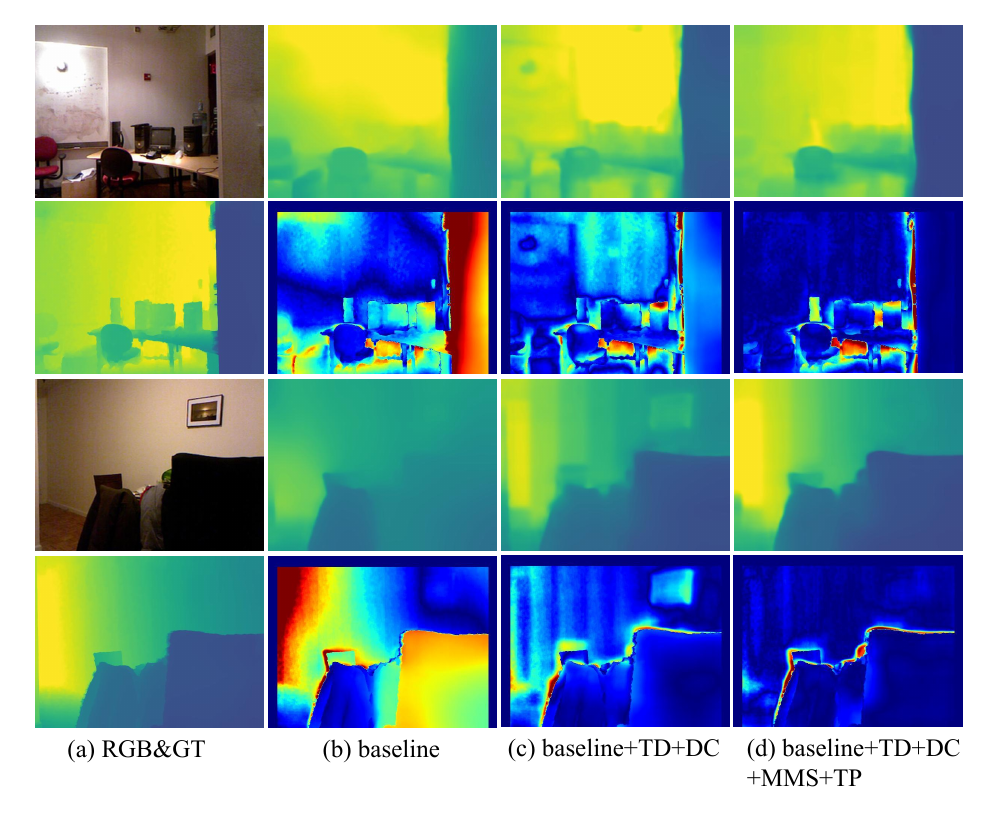}
    \caption{Visual results and error maps on the NYU dataset. Using ToF as DepthNet inputs (TD) and depth consistency loss (DC) can help recover better geometry. Furthermore, superior performance can be achieved with median of median scaling (MMS) and using ToF as PoseNet inputs (TP).}
    \label{fig:comparison2}
\end{figure}

In this section, we first describe the dataset preparation, implementation details, and evaluation metrics. Then, we provide results to validate our designs while building SelfToF. Lastly, we verify that our submanifold depth encoder and guided feature fusion can mitigate performance drops in various sparsity setups.

\subsection{Dataset Preparation}
As for the dataset preparation, we follow previous works \cite{jiang2021plnet} to sample one target frame every five frames and obtain training sets of size 47K from the NYU dataset \cite{silberman2012indoor}. The source frames are selected by sampling the $\pm 10$ frames around the target frame so that enough translation exists between source and target frames \cite{zhou2019moving}. To simulate the ToF signal from a lightweight ToF sensor, e.g., L5, we follow the process in DELTAR \cite{li2022deltar} to fit the histogram for each zone with a Gaussian distribution. For evaluation, we test on the official 654 samples with single-frame RGB input and paired ToF depth map. As for studies on the sparsity of ToF, we randomly fill some of the low-resolution depth maps with zeros. Notice that our SelfToF is scale-aware while previous self-supervised depth estimation is scale-ambiguous. Also, We conduct median scaling for the evaluation of scale-ambiguous methods.

\subsection{Implementation Details}
We implement our SelfToF with Pytorch \cite{paszke2019pytorch}. For training, we use the Adam optimizer \cite{kingma2014adam} with $\beta_1 = 0.9, \beta_2 = 0.999$. The learning rate is $1 \times e^{-4}$ for the first 30 epochs and remains $1 \times e^{-5}$ for the rest 10 epochs. The network is trained on four NVIDIA RTX 2080Ti GPUs with a batchsize of 8 on each GPU, which takes around 12 hours in total. The resolution of the RGB image and the depth map from a lightweight ToF sensor are $256 \times 256$ and $8 \times 8$, respectively. 

\subsection{Evaluation Metrics}

We follow previous works \cite{zhou2019moving,bhat2021adabins} to report standard metrics for depth prediction, including absolute relative error (abs\_rel), squared relative difference (sq\_rel), root mean squared error (rmse), rmse\_log, log\_{10} error and threshold accuracy ($\delta_i$).

\begin{table}[htbp]
\caption{Ablation of components in SelfToF on NYU dataset. Best performance among self-supervised methods is \textbf{bolded}.}
\resizebox{\columnwidth}{!}{
\centering
\begin{tabular}{|l|c|c|c|c|c|c|c|c|c|}
\hline
& SA & abs\_rel & sq\_rel & rmse & rmse\_log & log10 & a1   & a2   & a3   \\ \hline
baseline    & No    & 0.154            & 0.123           & 0.574        & 0.196             & 0.065         & 0.788        & 0.951        & 0.987        \\ 
+TD       & No     & 0.098            & 0.061           & 0.420        & 0.132             & 0.042         & 0.907        & 0.985        & 0.996        \\ 
+TD+DC  & No & 0.085            & 0.044           & 0.354        & 0.118             & 0.037         & 0.932        & 0.989        & \textbf{0.998}        \\ 
+TD+DC+MS & Yes & 0.078     & 0.041           & 0.343        & 0.121             & 0.035         & 0.933        & 0.986        & 0.996        \\ 
+TD+DC+MMS & Yes & 0.071     & 0.037           & 0.323        & 0.113             & 0.032         & 0.942        & 0.988        & 0.997        \\ 
+TD+DC+MMS+TP & Yes & \textbf{0.049}   & \textbf{0.029}           & \textbf{0.282}        & \textbf{0.091}             & \textbf{0.022}         & \textbf{0.960}        & \textbf{0.990}        & 0.997        \\ \hline
baseline+TD+S & Yes & 0.046  &     0.023       &   0.230     &     0.080        &  0.019     &  0.964    & 0.993   &   0.998  \\ \hline
\end{tabular}}
\label{table:ideal_ablate}
\end{table}

\subsection{Building SelfToF}

We start from a baseline self-supervised depth estimation framework, a single-scale MonoDepth2. Then, we add ToF as DepthNet inputs (TD), use depth consistency loss (DC), apply scale recovery (MS and MMS), and add ToF as PoseNet inputs (TP) sequentially, resulting in the final SelfToF. 

As listed in Table \ref{table:ideal_ablate}, adding TD and DC to the baseline can boost the performance, similar to previous self-supervised depth completion methods \cite{ma2019self,wong2020unsupervised}. Nevertheless, the results are still not scale-aware (SA) as we cannot set a large weight to $L_{dc}$. Consequently, we propose to find the actual scale by comparing the input ToF and predicted depth with MS and MMS. Both MS and MMS can recover the scale, yet MMS can lead to better numerical results since it exploits the zone-wise nature of lightweight ToF. Lastly, we find that using ToF as PoseNet inputs (TP) can additionally lead to a substantial performance increase. Overall, compared with the baseline method, we can reduce the abs\_rel and sq\_rel by 68.2\% and 76.4\%, respectively. Additionally, we train a baseline with TD and groundtruth supervision (S), and the result is in the last row. We can see only a small gap between SelfToF and this supervised model in most metrics.

Additionally, visual results on building SelfToF are given in Fig. \ref{fig:comparison2}. With TD and DC, the model can perform much better in areas where the photometric supervision is insufficient, e.g., low-light or over-exposed areas. Furthermore, an even greater performance boost can be realized with our proposed MMS and TP.

\subsection{Comparison with other methods}

We provide quantitative and qualitative results to examine the superior performance of SelfToF compared with other related methods, including MonoDepth2, nearest neighbour upsampling (NN), and guided filter (GF) \cite{he2012guided}.

As shown in Table \ref{table:comparison}, our SelfToF achieve the best performance among similar methods. Concretely, SelfToF outperform GF \cite{he2012guided} by reducing sq\_rel from 0.041 to 0.029, which is a 29.3\% reduction.

\begin{figure}[t]
    \centering
    \includegraphics[width=0.5\textwidth]{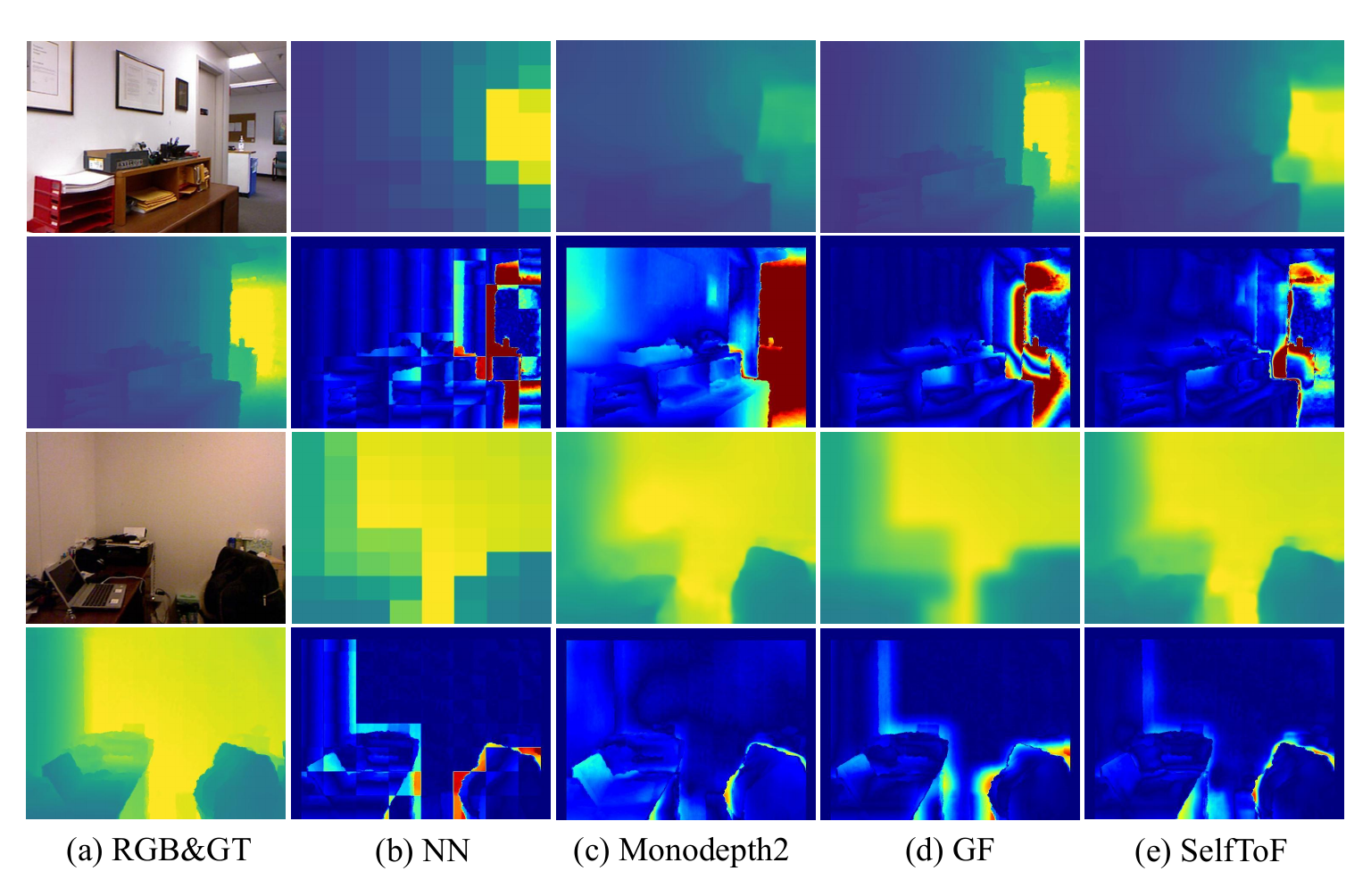}
    \caption{Visual results and error maps on NYU dataset. Compared with nearest upsampling (b), Monodepth2 (c), guided filter (d), our SelfToF (e) can produce depth maps with smaller errors and sharper object boundaries.}
    \label{fig:comparison1}
\end{figure}

We also visualize the results that different methods obtain in Fig. \ref{fig:comparison1}. Since we take advantage of self-supervised depth estimation, SelfToF can produce depth maps much sharper than nearest upsampling and guided filter \cite{he2012guided}. Additionally, SelfToF predicts more accurate output than MonoDepth2 as it exploits the depth information from ToF.

These quantitative and qualitative results indicate that SelfToF fully exploits the information from RGB images and depth maps from lightweight ToF sensors. Furthermore, SelfToF can output reasonable depth maps when sparsity in ToF exists. Traditional methods like NN and GF \cite{he2012guided} may fail in this case.

\begin{table}[htbp]
\caption{Quantitative comparison among SelfToF, Monodepth2, nearest neighbour upsampling (NN), and guided filter (GF).}
\resizebox{\columnwidth}{!}{
\centering
\begin{tabular}{|l|c|c|c|c|c|c|c|c|c|}
\hline
& SA & abs\_rel & sq\_rel & rmse & rmse\_log & log10 & a1   & a2   & a3   \\ \hline


MonoDepth2    & No    & 0.154            & 0.123           & 0.574        & 0.196             & 0.065         & 0.788        & 0.951        & 0.987        \\ \hline
NN    & Yes    & 0.070           & 0.064           & 0.405        & 0.140            & 0.031         & 0.926        & 0.970        & 0.988        \\ \hline

GF    & Yes    & 0.058            & 0.041           & 0.345        & 0.113             & 0.027         & 0.937        & 0.982        & 0.994        \\ \hline

SelfToF & Yes & \textbf{0.049}   & \textbf{0.029}           & \textbf{0.282}        & \textbf{0.091}             & \textbf{0.022}         & \textbf{0.960}        & \textbf{0.990}        & \textbf{0.997}       \\ \hline

\end{tabular}}
\label{table:comparison}
\end{table}

\subsection{Generalization on ScanNet}

Following prior works \cite{zhou2019moving,jiang2021plnet}, we evaluate the depth estimation and pose estimation on ScanNet \cite{dai2017scannet} with the NYU \cite{silberman2012indoor} trained model. The results are listed in Table \ref{table:scannet}. In addition to achieving much better depth-related results, SelfToF obtain smaller rotational error (rot) and translational angular error (tr). One may refer to SC-DepthV2 \cite{bian2021auto} or
MonoIndoor \cite{ji2021monoindoor} for even better pose estimation. Moreover, we find that SelfToF even generalize better than a supervised version of the baseline.

\begin{table}[htbp]
\caption{Generalization results on ScanNet\cite{dai2017scannet}.}
\resizebox{\columnwidth}{!}{
\centering
\begin{tabular}{|l|c|c|c|c|c|c|c|c|c|}
\hline
& SA & abs\_rel & sq\_rel & rmse & rmse\_log & log10 & a1 & rot(deg)   & tr(deg) \\ \hline
baseline & No & 0.181   & 0.109    & 0.439  & 0.225  & 0.076   & 0.724 & 1.740 & 32.770 \\ \hline
SelfToF & Yes & \textbf{0.044} & \textbf{0.015} & 0.169 & \textbf{0.084} & \textbf{0.020}  & \textbf{0.966} & \textbf{1.521}  & \textbf{28.217} \\ \hline
baseline+TD+S & Yes & 0.068   & 0.022    & \textbf{0.168}  & 0.101  & 0.027   & 0.933 & - & - \\ \hline
\end{tabular}}
\label{table:scannet}
\end{table}

\subsection{Handling the sparsity in ToF}

To investigate how sparsity in ToF influences the model performance, we train and test our SelfToF with ToF randomly missing. That is, we fill some zones with zeros randomly. 

As listed in Table \ref{table:missing}, when the sparsity rate (SR) increases, SelfToF obtain worse results on all metrics consistently. We argue that the zero-value zones originated from sparsity in the lightweight ToF sensor leads to wrong feature propagation during depth feature encoding. 

Consequently, we propose SelfToF* based on SelfToF by manipulating the depth feature encoding and propagation effectively. This upgrade is realized by using a submanifold depth encoder and a guided feature propagation module. By comparing the results from SelfToF* and SelfToF in Table \ref{table:missing}, we observe that SelfToF* achieve better performance in different sparsity settings.

\begin{table}[htbp]
\caption{Comparison between SelfToF and SelfToF* under different sparsity ratios (SR).}
\resizebox{\columnwidth}{!}{
\centering
\begin{tabular}{|l|c|c|c|c|c|c|c|c|c|}
\hline
& SR & abs\_rel & sq\_rel & rmse & rmse\_log & log10 & a1   & a2   & a3   \\ \hline
SelfToF    & 0.0    & 0.049   & 0.029         & 0.282        & 0.091             & 0.022         & 0.960        & 0.990        & 0.997        \\ \hline   
SelfToF    & 0.2    & 0.060            & 0.034           & 0.305        & 0.100             & 0.026 	        & 0.952          & 0.989        & 0.997        \\ \hline
SelfToF    & 0.4    & 0.070            & 0.042 	           & 0.337         & 0.111 	             & 0.031  	        & 0.939 	 	           & 0.987        & 0.997       \\ \hline
SelfToF*    & 0.0  & 0.049 	& 0.029 	& 0.281 	& 0.090 	& 0.022 	& 0.961 	& 0.990  &	0.997 \\ \hline
SelfToF*    & 0.2    & 0.053 	& 0.031 	& 0.291 &	0.095 	& 0.024 	& 0.957 	& 0.990 	& 0.997       \\ \hline
SelfToF*    & 0.4    & 0.061 	& 0.036 & 0.314 	& 0.104 	& 0.027 	& 0.949 	& 0.988 	& 0.997     \\ \hline
\end{tabular}}
\label{table:missing}
\end{table}

Furthermore, we provide more results to verify why using a Submanifold Depth Encoder (SDE) is necessary and how our Guided Feature Fusion (GFF) works.

First, looking at Table \ref{table:ablate_missing} where the results are obtained with 20\% sparsity, we observe that (1) using SDE does not lead to a performance boost. (2) GFF obtain much better results with SDE than used alone. These results indicate that a submanifold depth encoder is necessary to prevent improper feature propagation and allow better feature propagation in later stages. This differs from previous works \cite{uhrig2017sparsity,choi2021selfdeco} that use sparsity invariant convolution layers for only performance improvement.

\begin{table}[htbp]
\caption{Ablation studies of upgrading SelfToF to SelfToF*.}
\resizebox{\columnwidth}{!}{
\centering
\begin{tabular}{|l|c|c|c|c|c|c|c|c|}
\hline
& abs\_rel & sq\_rel & rmse & rmse\_log & log10 & a1   & a2   & a3   \\ \hline
SelfToF & 0.060 & 0.034 & 0.305 & 0.100 & 0.026 & 0.952 & 0.989 & \textbf{0.997} \\ \hline
+SDE & 0.060 & 0.034 & 0.304 & 0.100 & 0.026 & 0.953 & 0.989 & \textbf{0.997} \\ \hline
+GFF & 0.057 & 0.034 & 0.307 & 0.100 & 0.025 & 0.953 & 0.989 & \textbf{0.997} \\ \hline
+SDE+GFF & \textbf{0.053} & \textbf{0.031} & \textbf{0.291} & \textbf{0.095} & \textbf{0.024} & \textbf{0.957} & \textbf{0.990} & \textbf{0.997} \\ \hline
\end{tabular}}
\label{table:ablate_missing}
\end{table}



\begin{figure}[t]
    \centering
    \includegraphics[width=0.5\textwidth]{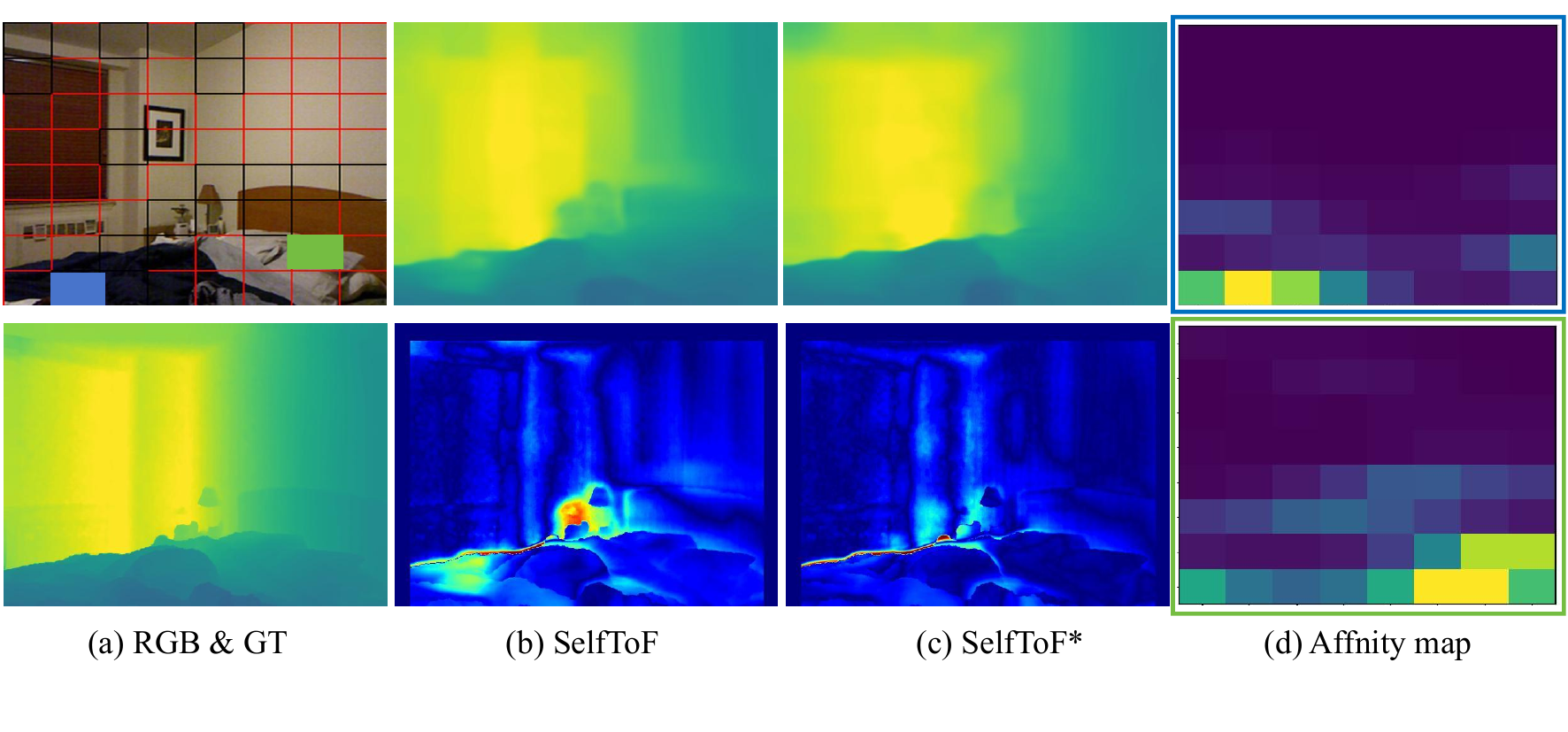}
    \caption{Visual results and error maps SelfToF and SelfToF* obtain. When some of the ToF is missing, SelfToF* can achieve lower errors in both valid zones and invalid zones. Also, the affinity map indicate that our guided feature fusion can indeed help propagate depth features correctly. }
    \label{fig:comparison_guided}
\end{figure}

Second, we visualize obtained affinity maps in Fig. \ref{fig:comparison_guided}. The zones with red boundaries are valid, while those with black boundaries are invalid. For the invalid blue zone, its associated affinity map or attention map is in the first row in Fig. \ref{fig:comparison_guided} (d). The similarity map concentrates at zones that belong to the same quilt. Similarly, the valid green zone has larger similarities with zones that are subject to the same pillow, as shown in the second row in Fig. \ref{fig:comparison_guided}. Additionally, SelfToF* achieves lower errors in these two regions. These results validate the effectiveness of our proposed feature
fusion and give a visual explanation of why it can help propagation features correctly.

Additionally, we study how the downsampling choice would affect the final performance. Specifically, given SelfToF with a submanifold depth encoder equipped, we ablate with regular convolution, depthwise convolution, and average pooling and present the results in Table \ref{table:ablate_ds}. We find that using average pooling can perform best, and fortunately, it is the most efficient one as no other parameters are required.

\begin{table}[htbp]
\caption{Ablation studies on the choice of downsampling strategies in guided feature fusion.}
\resizebox{\columnwidth}{!}{
\centering
\begin{tabular}{|l|c|c|c|c|c|c|c|c|}
\hline
Downsample Type & abs\_rel & sq\_rel & rmse & rmse\_log & log10 & a1   & a2   & a3   \\ \hline
Conv & 0.053 & 0.031 & 0.294 & 0.097 & 0.024 & 0.956 & 0.989 & \textbf{0.997} \\ \hline
Depthwise Conv & 0.055 & 0.032 & 0.296 & 0.096 & 0.024 & 0.956 & \textbf{0.990} & \textbf{0.997} \\ \hline
Avg Pooling & \textbf{0.053} & \textbf{0.031} & \textbf{0.291} & \textbf{0.095} & \textbf{0.024} & \textbf{0.957} & \textbf{0.990} & \textbf{0.997} \\ \hline

\end{tabular}}
\label{table:ablate_ds}
\end{table}

\section{CONCLUSIONS}

In this work, we propose SelfToF, a method that can enhance the low-resolution depth map from a lightweight ToF sensor with its paired RGB image. Furthermore, we build SelfToF* upon SelfToF by using a submanifold depth encoder and guided feature fusion to mitigate the performance drop due to the sparsity in ToF. 

For future work, one may consider collecting large-scale real-world datasets to facilitate studies on challenges rooted in real scenarios for lightweight ToF applications. Additionally, incorporating lightweight ToF sensors with depth foundation models \cite{yang2024depth,ke2023repurposing} to achieve generalized metric depth estimation results \cite{lin2024prompting} could be promising.

\bibliographystyle{IEEEtran}
\bibliography{IEEEfull}

\end{document}